\pdfoutput=1
\documentclass[11pt]{article}

\usepackage[final]{acl}

\usepackage{times}
\usepackage{latexsym}

\usepackage[T1]{fontenc}
\usepackage[utf8]{inputenc}

\usepackage{microtype}

\usepackage{inconsolata}

\usepackage{graphicx}

\usepackage{algorithmic,algorithm}
\renewcommand{\algorithmiccomment}[1]{\bgroup\hfill$\triangleright$~#1\egroup}
\usepackage[linguistics]{forest} 
\usepackage{synttree}
\usepackage{subcaption}
\usepackage{tikz-dependency}
\usepackage[cjk]{kotex}

\usepackage{amsmath}

\usepackage{booktabs}

\usepackage{langsci-gb4e}

\title{Non-Binary Bottom-Up Constituency Parsing Without Arity Actions
}

\author{
Jungyeul Park\raisebox{-0.1\height}{\includegraphics[height=0.7em]{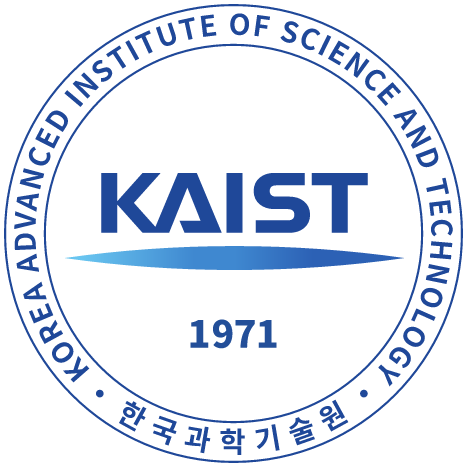}}\raisebox{-0.1\height}{\href{https://orcid.org/0000-0001-7828-6203}{
\includegraphics[height=0.7em]{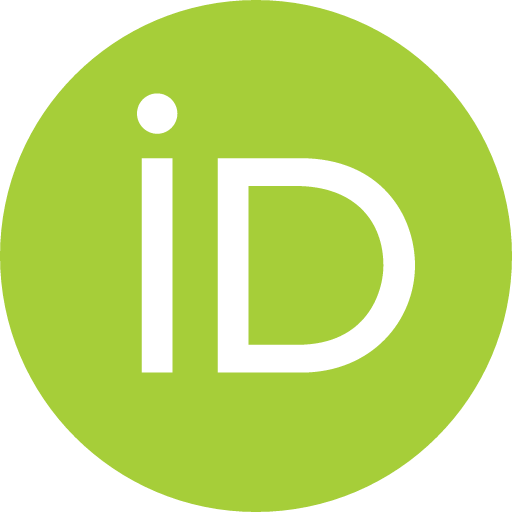}}}\thanks{Corresponding author: \url{jungyeul@kaist.ac.kr}}
Eunkyul Leah Jo\raisebox{-0.1\height}{\includegraphics[height=0.7em]{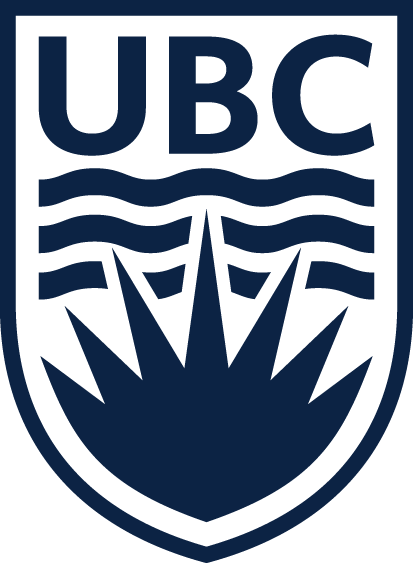}}\raisebox{-0.1\height}{\href{https://orcid.org/0009-0008-7893-5435}{
\includegraphics[height=0.7em]{ORCID_iD}}}~~~
Zihao Huang\raisebox{-0.1\height}{\includegraphics[height=0.7em]{UBC.png}}\raisebox{-0.1\height}{\href{https://orcid.org/0009-0003-4061-4031}{
\includegraphics[height=0.7em]{ORCID_iD}}}~~~
\\
\raisebox{-0.1\height}{\includegraphics[height=0.9em]{KAIST}} Korea Advanced Institute of Science \& Technology, Daejeon, South Korea\\
\raisebox{-0.1\height}{\includegraphics[height=0.9em]{UBC.png}} The University of British Columbia, Vancouver, Canada\\
\url{https://ct.kaist.ac.kr}~~~~~~~~ 
\url{https://linguistics.ubc.ca}\\}

\begin{document}

\maketitle

\begin{abstract}
Non-binary bottom-up constituency parsing commonly uses arity-specific reductions such as \(\textsc{Reduce-}X\#k\), which jointly specify the mother label and number of children. We show that arity need not be a primitive transition parameter. The proposed parser predicts the mother label separately and identifies the ordered child sequence from a delimiter-bounded stack configuration. In every well-formed reducible configuration, the active delimiter, label marker, and intervening completed items uniquely determine the reduction domain and hence its arity. This factorization removes label--arity-specific reduction actions while preserving direct construction of the original non-binary trees. Controlled experiments on PTB and CTB show that the delimiter-guided parser closely tracks an arity-specific baseline while using a substantially smaller action inventory. Its predicted trees retain the gold branching profile without systematic collapse at higher arities. The smaller inventory is obtained at the cost of longer derivations and boundary-sensitive errors involving delimiter insertion, preservation, and consumption.
\end{abstract}

\section{Introduction}

Transition-based constituency parsing builds phrase-structure trees through
local actions over a stack and an input buffer. A central distinction among
transition systems is how they make the boundary of an unfinished constituent
available before the constituent is completed. In top-down parsing, the boundary
is introduced by an open nonterminal \citep{dyer-etal-2016-recurrent}. In
in-order parsing, it is introduced by a projected nonterminal after the left
corner has been {constructed} \citep{liu-zhang-2017-order}. In purely
bottom-up parsing, no such marker is inherent to the derivation: the parent label
is introduced only after its children have already been recognized
\citep{gonzalez-rodriguez-2019-faster}.

This boundary problem is sharpened by non-binary parsing. Traditional bottom-up
systems avoid {variable-size reduction domains} by binarizing the treebank,
so that each reduction combines at most two stack items. Non-binary bottom-up
parsing instead constructs original treebank constituents directly, avoiding
the intermediate structure introduced by binarization. Existing systems achieve
this by encoding both the mother label and the number of children in a single
action, such as \(\textsc{Reduce-}X\#k\)
\citep{gonzalez-rodriguez-2019-faster}. In such a system, arity is an
action-level decision: \(k\) specifies how many completed stack items are
grouped under \(X\). The transition inventory is therefore tied to the observed
set of nonterminal--arity combinations.

We propose a non-binary bottom-up constituency parser in which arity is not an
explicit transition label. The parser separates label prediction from
{reduction-domain identification}. A constituent label is introduced by
\(\textsc{Nt}(X)\), while the reduction domain is recovered from delimiters
placed on the stack. The operations \(\textsc{D-Shift}\) and
\(\textsc{D-Reduce}\) make bottom-up constituent boundaries available without
requiring arity-specific reduction operators.

The resulting system retains direct construction of arbitrary-arity
constituents without binarization, but replaces arity classification with
{configuration-based domain recovery}. Its central property is arity
recoverability: in a well-formed delimiter-bounded reduction configuration, the
number of children is uniquely determined by the completed stack items between
the active delimiter and the label marker. Arity is thus {a derived property
of the reducible configuration rather than a primitive action parameter}.

This factorization gives bottom-up non-binary parsing a different structural
primitive. Existing arity-specific systems identify the reduction domain by
specifying its size in the action label. Our system makes the domain recoverable
from the parser configuration. The contribution is therefore not only a smaller
action inventory, but a reformulation of bottom-up reduction in which constituent
labels and reduction domains are represented independently.

{Controlled experiments on PTB and CTB show that the delimiter-guided parser
closely tracks its arity-specific counterpart while using substantially fewer
action types. Its predicted trees retain the observed non-binary arity profile
without systematic degradation at higher arities. The factorization instead
introduces a different operational cost: longer derivations and errors involving
delimiter insertion, preservation, and consumption. These results support the
central representational claim that direct non-binary bottom-up construction
requires an identifiable reduction domain, but not an arity-indexed reduction
action.}

\section{Background}
\label{sec:background}

{Transition-based constituency parsers differ not only in construction order
but also in where they represent the information identifying an unfinished
constituent. The unresolved problem for purely bottom-up non-binary parsing is
how to identify a variable-size reduction domain when no prior nonterminal
marker supplies its left boundary.}

\subsection{Transition-based constituency parsing}

{Transition-based constituency parsing constructs phrase-structure trees
through local actions over a stack and an input buffer. Early bottom-up systems
shift terminals onto the stack and apply unary or binary reductions, usually
over binarized trees
\citep{sagae-lavie-2005-classifier,wang-etal-2006-fast,zhang-clark-2009-transition,zhu-etal-2013-fast,watanabe-sumita-2015-transition,liu-zhang-2017-shift}. Non-binary constituents
are consequently represented by sequences of binary structures and recovered
only through debinarization.}

{Direct non-binary systems instead construct the annotated treebank
constituents. In top-down parsing, an open nonterminal introduced by
\(\textsc{Nt}(X)\) supplies the mother label and fixes the left boundary before
the children are constructed \citep{dyer-etal-2016-recurrent}. A subsequent
\(\textsc{Reduce}\) constructs the constituent over the completed children above
that marker.}

{In-order parsing introduces the mother label after the left corner has been
constructed \citep{liu-zhang-2017-order}. Its reduction domain consists of the
completed left corner immediately below the marker and the completed children
above it. Top-down and in-order parsing therefore recover the child sequence
from an explicit marker configuration rather than from an arity-specific
reduction action.}

{Purely bottom-up non-binary parsing introduces the mother label only after
all children have been constructed. The label cannot therefore serve as a prior
left-boundary marker. The system of
\citet{gonzalez-rodriguez-2019-faster} resolves this problem by encoding the
number of children directly in the reduction action.}

\subsection{Non-binary bottom-up parsing}

{Traditional bottom-up parsers avoid variable-size reduction domains by
binarizing the treebank. A constituent with more than two children is replaced
by a sequence of binary nodes and constructed through successive binary
reductions. Headed variants additionally require a head-assignment procedure,
and every predicted tree must be debinarized before evaluation against the
original treebank representation. The parser is thus trained and decoded in a
refined representation rather than directly in the target tree space.}

{The non-binary bottom-up parser of
\citet{gonzalez-rodriguez-2019-faster} removes this intermediate representation.
Its action \(\textsc{Reduce-}X\#k\) pops the \(k\) topmost completed stack items,
constructs a constituent labeled \(X\) over their ordered sequence, and pushes
the result onto the stack. Unary, binary, and higher-arity constituents are
therefore constructed directly in the original non-binary representation.}

{This design makes every observed label--arity combination a distinct
classifier target. The transition inventory consequently grows with the
observed combinations, and the reduction domain is identified by predicting its
size \(k\). Arity is thus encoded as a primitive property of the action even
though it is also the cardinality of the selected child sequence.}

{Our system retains direct non-binary bottom-up construction while changing
where the reduction-domain information is represented. The mother label is
predicted separately, and the ordered child sequence is recovered from a
delimiter-bounded stack suffix. Once that suffix is uniquely identified, arity
follows from the number of completed children it contains. The central contrast
is therefore between arity encoded in an action label and arity derived from a
reducible parser configuration.}

\section{Bottom-up parsing with delimiter-guided reduction}
\label{sec:method}

We define a non-binary bottom-up transition system in which reduction arity is derived from the parser configuration rather than encoded in an action. The mother label is introduced by \(\textsc{Nt}(X)\), and the ordered child sequence is identified by the nearest active delimiter on the stack. Phrase-level construction therefore separates label prediction from reduction-domain identification.

Because a bottom-up parser introduces a mother only after constructing its children, the mother marker cannot delimit their left boundary in advance. We use a delimiter \(D\), implemented as the empty label marker \(\textsc{Nt}()\), to record that boundary. \(\textsc{d-Shift}\) introduces \(D\), and \(\textsc{d-Reduce}\) preserves it when a higher constituent shares the same left boundary. Delimiters are auxiliary control symbols and do not occur in the output tree.

\subsection{Parser configurations}

A parser configuration is a triple \((\sigma,i,f)\), where \(\sigma\) is the stack, \(i\) is the index of the next input token, and \(f\) is a completion flag. For a zero-indexed sentence \(w=w_0\cdots w_{n-1}\), the initial configuration is
\[
(\epsilon,0,\texttt{false}),
\]
and a final configuration has the form
\[
(T,n,\texttt{true}),
\]
where \(T\) is a single completed root constituent. Every transition other than \(\textsc{Finish}\) requires \(f=\texttt{false}\).

The stack contains terminals, completed subtrees, label markers, and delimiters. Stack concatenation is written as \(\sigma\mid x\), and a completed constituent with label \(X\) and ordered children \(s_1,\ldots,s_k\) is written as \(X(s_1,\ldots,s_k)\). We identify a label marker with its label \(X\). A delimiter \(D\) marks the left boundary of an active reduction domain but is neither a child nor a node of the resulting tree.

\subsection{Transition inventory}

The transition inventory consists of \(\textsc{Shift}\), \(\textsc{d-Shift}\), \(\textsc{Nt}(X)\), \(\textsc{u-Reduce}\), \(\textsc{Reduce}\), \(\textsc{d-Reduce}\), and \(\textsc{Finish}\). The two shift actions consume input, \(\textsc{Nt}(X)\) introduces a constituent label, and the three reduction actions perform preterminal or phrase-level composition. The structural definitions are given in Figure~\ref{fig:transitions}.

The shift actions require \(i<n\). In a phrase-level reduction, \(k\geq 1\), each \(s_j\) is a completed subtree, \(X\) is the topmost label marker, and \(D\) is the nearest delimiter below \(X\). Consequently, \(s_1,\ldots,s_k\) constitute the complete ordered sequence between \(D\) and \(X\).

\begin{figure*}[!th]
\centering
\small
\[
\begin{array}{ll}
\textsc{Shift}: &
  (\sigma,i,\texttt{false})
  \Rightarrow
  (\sigma\mid w_i,i+1,\texttt{false})
\\[1ex]

\textsc{d-Shift}: &
  (\sigma,i,\texttt{false})
  \Rightarrow
  (\sigma\mid D\mid w_i,i+1,\texttt{false})
\\[1ex]

\textsc{Nt}(X): &
  (\sigma,i,\texttt{false})
  \Rightarrow
  (\sigma\mid X,i,\texttt{false})
\\[1ex]

\textsc{u-Reduce}: &
  (\sigma\mid w\mid X,i,\texttt{false})
  \Rightarrow
  (\sigma\mid X(w),i,\texttt{false})
\\[1ex]

\textsc{Reduce}: &
  (\sigma\mid D\mid s_1\mid\cdots\mid s_k\mid X,i,\texttt{false})
  \Rightarrow
  (\sigma\mid X(s_1,\ldots,s_k),i,\texttt{false})
\\[1ex]

\textsc{d-Reduce}: &
  (\sigma\mid D\mid s_1\mid\cdots\mid s_k\mid X,i,\texttt{false})
  \Rightarrow
  (\sigma\mid D\mid X(s_1,\ldots,s_k),i,\texttt{false})
\\[1ex]

\textsc{Finish}: &
  (T,n,\texttt{false})
  \Rightarrow
  (T,n,\texttt{true}) .
\end{array}
\]
\caption{Delimiter-guided bottom-up transitions. \textsc{u-Reduce} constructs a preterminal; \textsc{Reduce} and \textsc{d-Reduce} compose the delimiter-bounded child sequence, respectively consuming and preserving \(D\).}
\label{fig:transitions}
\end{figure*}

\subsection{\textsc{Shift}, \textsc{d-Shift}, and boundary insertion}

\(\textsc{Shift}\) pushes the next input token onto the stack, whereas \(\textsc{d-Shift}\) first inserts a delimiter:
\[
\begin{aligned}
\textsc{Shift}:&\quad
\sigma \Rightarrow \sigma\mid w_i,\\
\textsc{d-Shift}:&\quad
\sigma \Rightarrow \sigma\mid D\mid w_i.
\end{aligned}
\]
The latter marks \(w_i\) as the leftmost terminal of a new phrase-level reduction domain.

Because terminals are immediately dominated by part-of-speech labels, the grandparent of a terminal is its nearest phrase-level projection. Let \(g(w_i)\) denote the grandparent of \(w_i\), and let \(d(w_i)\) denote its depth in the gold tree, with greater depth corresponding to a lower tree position. The static oracle selects \(\textsc{d-Shift}\) at \(w_i\) according to
% \[
% \textsc{d-Shift}(w_i)
% =
% \begin{cases}
% 1,
%   & \text{if } i=0,\\
% 1,
%   & \text{if } g(w_i)\neq g(w_{i-1})
%     \text{ and } d(w_i)\geq d(w_{i-1}),\\
% 0,
%   & \text{otherwise}.
% \end{cases}
% \]
\[
\textsc{d-Shift}(w_i)
=
\begin{cases}
1,
  & \text{if } i=0,\\
1,
  &
  \begin{aligned}[t]
  \text{if }& g(w_i)\neq g(w_{i-1})\\
             & \text{and } d(w_i)\geq d(w_{i-1}),
  \end{aligned}\\
0,
  & \text{otherwise}.
\end{cases}
\]
The depth condition admits boundaries between distinct phrase-level projections at the same depth, including the boundary between a subject noun phrase and a following verbal phrase. The initial delimiter supplies the left boundary of the root domain without requiring a separate stack-bottom symbol.

The oracle associates a delimiter inserted before \(w_i\) with the highest gold phrasal ancestor of \(w_i\) whose left boundary is \(w_i\). The delimiter remains active while constituents properly contained in that domain are constructed and is consumed when the associated constituent is completed.

\subsection{\textsc{Nt} and label prediction}

\(\textsc{Nt}(X)\) pushes the marker \(X\) after the children of the intended constituent have been shifted or constructed. Unlike an open nonterminal in top-down parsing, it does not establish the left boundary of a future constituent. It supplies only the mother label; the active delimiter identifies the ordered child sequence.

The same action introduces part-of-speech and phrase labels. For a preterminal, \(\textsc{Nt}(X)\) follows its terminal and is discharged by \(\textsc{u-Reduce}\). For a phrase-level constituent, it follows the completed children and is discharged by \(\textsc{Reduce}\) or \(\textsc{d-Reduce}\). The mode of composition is therefore determined by the reduction action, whereas \(X\) determines the label of the constructed node.

\subsection{\textsc{u-Reduce}, \textsc{Reduce}, \textsc{d-Reduce}, and domain recovery}

\(\textsc{u-Reduce}\) is restricted to preterminal construction:
\[
\sigma\mid w\mid X
\Rightarrow
\sigma\mid X(w),
\]
where \(w\) is a terminal and \(X\) is its part-of-speech label. It is not a general unary phrase-level reduction and does not depend on a delimiter.

For phrase-level construction, the nearest active delimiter and the topmost label marker delimit a unique configuration
\[
\sigma\mid D\mid s_1\mid\cdots\mid s_k\mid X.
\]
Both phrase-level reduction actions construct \(X(s_1,\ldots,s_k)\). The difference concerns only the status of \(D\): \(\textsc{Reduce}\) consumes it, whereas \(\textsc{d-Reduce}\) leaves it on the stack for a higher constituent with the same left boundary.

Let \(a(D)\) be the gold constituent associated by the oracle with \(D\). If the current reduction constructs \(C\), the oracle selects
\[
\begin{array}{ll}
\textsc{d-Reduce},
  & \text{if } C \text{ is a proper descendant of } a(D),\\
\textsc{Reduce},
  & \text{if } C=a(D).
\end{array}
\]
Thus, \(\textsc{d-Reduce}\) completes a constituent without closing the active domain, while \(\textsc{Reduce}\) completes the constituent to which the delimiter is assigned.

\paragraph{Arity recoverability}

In every well-formed phrase-level reduction configuration
\[
\sigma\mid D\mid s_1\mid\cdots\mid s_k\mid X,
\]
the arity of the constructed constituent is uniquely determined as the number of completed subtrees strictly between the active delimiter \(D\) and the label marker \(X\). A smaller value would omit a subtree within the domain, whereas a larger value would cross its left boundary. Hence neither \(\textsc{Reduce}\) nor \(\textsc{d-Reduce}\) requires an arity parameter.

The distinction among the reduction actions is independent of phrase-level arity. \(\textsc{u-Reduce}\) constructs a preterminal, \(\textsc{Reduce}\) closes an active phrase-level domain, and \(\textsc{d-Reduce}\) preserves that domain after composition. Unary, binary, and higher-arity phrase-level constituents are all constructed by the same delimiter-guided schema.

\paragraph{Relation to arity-specific reduction}

For every applicable delimiter-guided phrase-level reduction, the active delimiter and label marker determine a unique child sequence \(s_1,\ldots,s_k\). After auxiliary control symbols are disregarded, the corresponding arity-specific action is \(\textsc{Reduce-}X\#k\): both operations construct the same mother \(X\) over the same ordered children.

Conversely, an arity-specific reduction has the same local tree construction whenever a reachable delimiter-guided configuration contains an active delimiter immediately before the corresponding \(k\) children. The delimiter must have been introduced by a valid \(\textsc{d-Shift}\); it cannot be inserted arbitrarily at reduction time. The local correspondence therefore concerns the representation of an already identified reduction domain: the arity-specific action records its cardinality in the action label, whereas the delimiter-guided system recovers that cardinality from the reducible configuration.

\subsection{Static-oracle example}
\label{app:example}

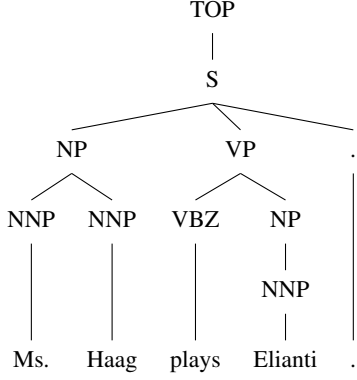
\begin{figure}
\centering
\small
\begin{forest}
for tree={
  align=center,
  rounded corners,
  if n children=0{
    tier=word
  }{}
}
[TOP
  [S
    [NP
      [NNP [Ms.]]
      [NNP [Haag]]
    ]
    [VP
      [VBZ [plays]]
      [NP
        [NNP [Elianti]]
      ]
    ]
    [. [.]]
  ]
]
\end{forest}
\caption{Example PTB-style constituency tree.}
\label{fig:oracle-example-tree}
\end{figure}

Figure~\ref{fig:oracle-example-tree} shows the PTB-style tree used to illustrate the static oracle. Its phrase-level branching comprises a binary subject noun phrase, a unary object noun phrase, a binary verbal phrase, a ternary sentence constituent, and a unary root.

Table~\ref{tab:oracle-example} gives the corresponding oracle derivation. The input column records the terminal consumed by a shift action; terminals are not parameters of the transition label.

\begin{table}
\centering
\small
\begin{tabular}{rll}
\toprule
Step & Transition & Input \\
\midrule
1  & \textsc{d-Shift} & \textit{Ms.} \\
2  & \(\textsc{Nt}(\mathrm{NNP})\) & \\
3  & \textsc{u-Reduce} & \\
4  & \textsc{Shift} & \textit{Haag} \\
5  & \(\textsc{Nt}(\mathrm{NNP})\) & \\
6  & \textsc{u-Reduce} & \\
7  & \(\textsc{Nt}(\mathrm{NP})\) & \\
8  & \textsc{d-Reduce} & \\
9  & \textsc{d-Shift} & \textit{plays} \\
10 & \(\textsc{Nt}(\mathrm{VBZ})\) & \\
11 & \textsc{u-Reduce} & \\
12 & \textsc{d-Shift} & \textit{Elianti} \\
13 & \(\textsc{Nt}(\mathrm{NNP})\) & \\
14 & \textsc{u-Reduce} & \\
15 & \(\textsc{Nt}(\mathrm{NP})\) & \\
16 & \textsc{Reduce} & \\
17 & \(\textsc{Nt}(\mathrm{VP})\) & \\
18 & \textsc{Reduce} & \\
19 & \textsc{Shift} & \textit{.} \\
20 & \(\textsc{Nt}(\texttt{.})\) & \\
21 & \textsc{u-Reduce} & \\
22 & \(\textsc{Nt}(\mathrm{S})\) & \\
23 & \textsc{d-Reduce} & \\
24 & \(\textsc{Nt}(\mathrm{TOP})\) & \\
25 & \textsc{Reduce} & \\
26 & \textsc{Finish} & \\
\bottomrule
\end{tabular}
\caption{Static-oracle derivation for the tree in Figure~\ref{fig:oracle-example-tree}.}
\label{tab:oracle-example}
\end{table}

The initial \textsc{d-Shift} introduces the delimiter associated with \(\mathrm{TOP}\), the highest phrasal ancestor whose left boundary is \textit{Ms.} The following \textsc{Shift} consumes \textit{Haag} without introducing another delimiter because the two terminals belong to the same subject noun-phrase domain. Once their preterminals have been constructed, \(\textsc{Nt}(\mathrm{NP})\) and \textsc{d-Reduce} construct the subject noun phrase while preserving the initial delimiter.

The delimiters introduced before \textit{plays} and \textit{Elianti} are associated with the verbal phrase and object noun phrase, respectively. The \textsc{Reduce} at step~16 constructs the object noun phrase and consumes its delimiter. The \textsc{Reduce} at step~18 then constructs the verbal phrase and consumes the delimiter introduced before \textit{plays}.

The punctuation is shifted without a new delimiter. After its preterminal has been constructed, \(\textsc{Nt}(\mathrm{S})\) and \textsc{d-Reduce} construct the sentence constituent while preserving the initial delimiter, since \(\mathrm{S}\) is a proper descendant of \(\mathrm{TOP}\). Finally, \(\textsc{Nt}(\mathrm{TOP})\) and \textsc{Reduce} construct the root and consume that delimiter. \textsc{Finish} applies after the input is exhausted and the stack contains the single completed root.

The unary, binary, and ternary phrase-level reductions in this derivation use the same action schema. Their arities are determined by the completed subtrees between the active delimiter and the label marker.

\section{Experiments and results}
\label{sec:experiments}

We test whether arity must be encoded explicitly in direct non-binary bottom-up reduction. The primary comparison is between an arity-specific parser with actions of the form \(\textsc{Reduce-}X\#k\) and a delimiter-guided parser that identifies the complete reduction domain from its configuration and derives arity from the resulting child sequence. A binary bottom-up parser trained on binarized trees provides a reference for the alternative binary representation. Complete transition definitions and static-oracle derivations for the reimplemented baselines are provided in Appendix~\ref{app:previous-work-reimplementation}.

Experiments are conducted on the Wall Street Journal portion of the Penn Treebank (PTB), using Sections~02--21 for training, Section~22 for development, and Section~23 for testing \citep{marcus-etal-1993-building}, and on Chinese Treebank 5.1 (CTB) using the standard split from prior work \citep{xue-etal-2005-ctb}. Preprocessing and training details are given in Appendix~\ref{app:experimental-details}.

All three systems use the same input representations, neural architecture, optimization, decoding, and evaluation procedures. They differ only in their transition inventories, structural applicability conditions, and static oracles. The arity-specific system reimplements \citet{gonzalez-rodriguez-2019-faster}; the delimiter-guided system predicts the mother label separately and uses delimiter-sensitive shift and reduction actions whose reduction operators have neither label nor arity parameters. The binary system uses unary and headed binary reductions \citep{sagae-lavie-2005-classifier} and is evaluated in the original treebank representation after debinarization.

Table~\ref{tab:main-results} reports development-set labeled \(F_1\) using \textsc{evalb}\footnote{\url{https://nlp.cs.nyu.edu/evalb/}} and \textsc{jp-evalb}. \textsc{evalb} follows the conventional punctuation-excluding evaluation \citep{black-etal-1991-procedure}, whereas \textsc{jp-evalb} retains punctuation in constituent evaluation \citep{jo-etal-2024-novel,park-etal-2024-jp}. On PTB, the delimiter-guided and arity-specific scores differ by at most 0.08 points; on CTB, the delimiter-guided scores are lower by 0.68 and 0.57 points under \textsc{evalb} and \textsc{jp-evalb}, respectively. These differences show that removing arity-indexed actions does not produce a substantial loss in direct non-binary parsing accuracy. The binary parser obtains higher scores, but because it is trained on binarized trees, the difference cannot be attributed specifically to explicit arity actions.

\begin{table}
\centering
\small
\begin{tabular}{c cccc}
\toprule
 & \multicolumn{2}{c}{PTB} & \multicolumn{2}{c}{CTB} \\
\cmidrule(lr){2-3}\cmidrule(lr){4-5}
 & \textsc{evalb} & \textsc{jp-evalb}
 & \textsc{evalb} & \textsc{jp-evalb} \\
\midrule
\(\mathcal{B}\) & 94.52 & 93.77 & 95.70 & 94.07 \\
\(\mathcal{A}\) & 93.04 & 92.37 & 93.55 & 91.86 \\
\(\mathcal{D}\) & 92.96 & 92.31 & 92.87 & 91.29 \\
\bottomrule
\end{tabular}
\caption{{Development-set labeled \(F_1\) on PTB and CTB. \(\mathcal{B}\), \(\mathcal{A}\), and \(\mathcal{D}\) denote the binary, arity-specific non-binary, and delimiter-guided non-binary bottom-up systems, respectively.}}
\label{tab:main-results}
\end{table}

{Table~\ref{tab:previous-work} reports test-set \textsc{evalb} and \textsc{jp-evalb} scores together with representative published results. The published systems provide external reference points because their architectures, preprocessing, and training conditions differ. The controlled comparison is between the arity-specific and delimiter-guided systems implemented within the present framework.}

% {All current-work results include every test sentence and are computed from predictions whose terminal yields have been validated against the corresponding gold trees. The corrected arity-specific implementation eliminates the historical batch-alignment corruption; the remaining PTB decode that reaches the transition limit is represented by a yield-preserving fallback and retained as a parser failure in the evaluation. No sentence is discarded.}

{Under this controlled evaluation, the delimiter-guided parser remains competitive with the arity-specific parser despite removing \(\textsc{Reduce-}X\#k\). The result supports the representational claim examined in Section~\ref{sec:analysis}: bottom-up non-binary reduction requires an identifiable reduction domain, but its arity need not be predicted as an independent action parameter.}

\begin{table*}
    \centering
    %\footnotesize
\resizebox{\textwidth}{!}{    
    \begin{tabular}{r cccc l}
    \toprule
     & \multicolumn{2}{c}{PTB} & \multicolumn{2}{c}{CTB} & \\
\cmidrule(lr){2-3}\cmidrule(lr){4-5}
 & \multicolumn{2}{c}{\textsc{evalb}} & \multicolumn{2}{c}{\textsc{evalb}} &\\
\midrule
\citet{watanabe-sumita-2015-transition} 
& \multicolumn{2}{c}{90.68} 
& \multicolumn{2}{c}{84.33} 
& Binary bottom-up  \\

\citet{liu-zhang-2017-shift} 
& \multicolumn{2}{c}{91.70} 
& \multicolumn{2}{c}{85.50} 
& Binary bottom-up  \\

\citet{dyer-etal-2016-recurrent}
& \multicolumn{2}{c}{92.40} 
& \multicolumn{2}{c}{82.70} 
& Non-binary top-down\\

\citet{liu-zhang-2017-order}
& \multicolumn{2}{c}{91.80} 
& \multicolumn{2}{c}{86.10} 
& Non-binary in-order\\

\citet{gonzalez-rodriguez-2019-faster}
& \multicolumn{2}{c}{91.70} 
& \multicolumn{2}{c}{86.80} 
& Non-binary arity-specific bottom-up\\
\midrule
 & \textsc{evalb} & \textsc{jp-evalb} & \textsc{evalb} & \textsc{jp-evalb} &\\
\cmidrule(lr){2-3}\cmidrule(lr){4-5}
Current work $\mathcal{B}$ & 94.11 & 93.64 & 92.81 & 91.08 & Binary bottom-up \\
$\mathcal{A}$ & 92.45 & 91.89 & 91.02 & 89.27 & Non-binary arity-specific bottom-up \\
$\mathcal{D}$ & 92.90 & 92.33 & 90.90 & 89.13 & Non-binary delimiter-guided bottom-up \\
\bottomrule
    \end{tabular}
}
\caption{Test F1 on PTB and CTB with \textsc{evalb} and \textsc{jp-evalb}. Published prior results are shown for reference; the current-work rows compare the arity-specific and delimiter-guided parsers under the same implementation framework.}
\label{tab:previous-work}
\end{table*}

\section{Analysis and discussion}
\label{sec:analysis}

{The empirical comparison isolates two representations of a bottom-up reduction domain. The arity-specific parser encodes the domain size in actions of the form \(\textsc{Reduce-}X\#k\). The delimiter-guided parser instead represents the domain in its configuration: \(\textsc{Nt}(X)\) supplies the mother label, and the suffix bounded by the active delimiter supplies the ordered child sequence. Arity is therefore a derived property of a well-formed configuration rather than a primitive action parameter.}

\subsection{Action inventory factorization}

{Figure~\ref{fig:inventory-composition} summarizes the action schemas of the three systems. Binary bottom-up parsing fixes arity through unary and binary reduction types. Arity-specific non-binary parsing requires a distinct reduction action for each observed label--arity combination. Delimiter-guided parsing predicts the mother label separately, while \(\textsc{Reduce}\) and \(\textsc{D-Reduce}\) are independent of both label and arity.}

{Table~\ref{tab:action-inventory} shows the resulting difference in inventory size. Delimiter guidance reduces the arity-specific inventory from 202 to 78 actions on PTB and from 190 to 64 actions on CTB without changing the non-binary target representation. The reduction follows from eliminating the cross-product between mother labels and constituent arities.}

\begin{figure*}
\centering
\small
\begin{tabular}{c l}
\toprule
 & {Action schemas} \\
\midrule
\(\mathcal{B}\)
&
{\(\textsc{Shift},\ \textsc{Reduce-Unary-}X,\ 
\textsc{Reduce-Left-}X,\ \textsc{Reduce-Right-}X,\ 
\textsc{Finish}\)}
\\[0.5ex]

\(\mathcal{A}\)
&
{\(\textsc{Shift},\ \textsc{Reduce-}X\#k,\ 
\textsc{Finish}\)}
\\[0.5ex]

\(\mathcal{D}\)
&
{\(\textsc{Shift},\ \textsc{D-Shift},\ \textsc{Nt}(X),\ 
\textsc{U-Reduce},\ \textsc{Reduce},\ \textsc{D-Reduce},\ 
\textsc{Finish}\)}
\\
\bottomrule
\end{tabular}
\caption{{Action schemas of the binary, arity-specific non-binary, and delimiter-guided bottom-up systems.}}
\label{fig:inventory-composition}
\end{figure*}

\begin{table}
\centering
\small
\begin{tabular}{ccc}
\toprule
 & PTB & CTB \\
\midrule
\(\mathcal{B}\) & 126 & 148 \\
\(\mathcal{A}\) & 202 & 190 \\
\(\mathcal{D}\) & 78  & 64  \\
\bottomrule
\end{tabular}
\caption{{Action-inventory sizes of the three bottom-up systems.}}
\label{tab:action-inventory}
\end{table}

\subsection{Reduction arity distribution}

{The arity distribution tests whether configuration-based domain recovery changes the branching profile of the predicted trees. Systematic boundary errors could bias the parser toward shorter reduction domains even though its transition inventory imposes no arity bound. Arity is measured as the number of immediate children, including punctuation represented as a constituent child.}

{Table~\ref{tab:arity-distribution} shows that the delimiter-guided predictions track the gold distributions across unary, binary, ternary, and higher-arity constituents. The parser therefore does not collapse toward binary or low-arity outputs when label--arity-specific actions are removed. The observed branching profile is recovered from delimiter-bounded configurations rather than encoded directly in the action inventory.}

\begin{table}
\centering
\footnotesize
\begin{tabular}{r ccccc}
\toprule
 & \(k=1\) & \(k=2\) & \(k=3\) & \(k=4\) & \(k\geq5\) \\
\midrule
PTB {Gold} & 8879 & 23840 & 7381 & 2317 & 1859 \\
\(\mathcal{A}\) & 8660 & 23706 & 7331 & 2269 & 1950 \\
\(\mathcal{D}\) & 8696 & 23697 & 7403 & 2307 & 1897 \\
\midrule
CTB {Gold} & 3482 & 3698 & 1121 & 184 & 235 \\
\(\mathcal{A}\) & 3424 & 3736 & 1137 & 177 & 222 \\
\(\mathcal{D}\) & 3426 & 3706 & 1081 & 193 & 238 \\
\bottomrule
\end{tabular}
\caption{{Immediate-child arity distributions in the gold trees and system outputs.}}
\label{tab:arity-distribution}
\end{table}

\subsection{High-arity constituent F-score}

{Higher-arity constituents provide a direct test of configuration-based domain recovery because an incorrect boundary changes both the child sequence and its derived arity. The arity-specific parser selects the child count directly through its reduction action, whereas the delimiter-guided parser must recover the complete sequence from the bounded stack suffix.}

\begin{table}
\centering
\footnotesize
\begin{tabular}{r ccccc}
\toprule
 & \(k=1\) & \(k=2\) & \(k=3\) & \(k=4\) & \(k\geq5\) \\
\midrule
PTB
\(\mathcal{A}\) & 92.64 & 90.05 & 85.79 & 78.02 & 83.01 \\
\(\mathcal{D}\) & 92.77 & 90.52 & 86.30 & 79.02 & 83.33 \\
\midrule
CTB
\(\mathcal{A}\) & 92.90 & 88.30 & 82.46 & 69.25 & 74.84 \\
\(\mathcal{D}\) & 93.31 & 88.20 & 81.83 & 68.44 & 72.73 \\
\bottomrule
\end{tabular}
\caption{{Labeled constituent \(F_1\) by immediate-child arity.}}
\label{tab:arity-fscore}
\end{table}

{Table~\ref{tab:arity-fscore} shows no collapse at larger arities. The delimiter-guided system closely tracks the arity-specific system across the branching range, although the direction and magnitude of the difference vary between treebanks. The structural implication is that higher-arity constituents can be constructed without assigning each arity a separate reduction type. Once the reduction domain is uniquely identified, its child count follows from the configuration.}

\subsection{Transition sequence length}
\label{sec:transition-length}

{Action factorization changes derivation length. Arity-specific parsing combines mother-label prediction, domain-size selection, and constituent construction in one reduction. Delimiter-guided parsing separates label introduction from the arity-independent reduction. The boundary-bearing \(\textsc{D-Shift}\) replaces an ordinary \(\textsc{Shift}\) at selected terminals rather than adding a separate transition. Table~\ref{tab:transition-length} reports the resulting average sequence lengths.}

\begin{table}
\centering
\small
\begin{tabular}{c cc}
\toprule
 & PTB & CTB \\
\midrule
\(\mathcal{B}\) & 47.70 & 54.64 \\
\(\mathcal{A}\) & 43.47 & 59.45 \\
\(\mathcal{D}\) & 56.76 & 77.56 \\
\bottomrule
\end{tabular}
\caption{{Average transition-sequence length per sentence under the three bottom-up representations.}}
\label{tab:transition-length}
\end{table}

{The delimiter-guided representation exchanges a smaller action inventory for a longer derivation. These measurements characterize complementary aspects of the state--action factorization: inventory size measures the number of prediction types, whereas derivation length measures the number of sequential decisions. Neither quantity alone determines computational efficiency.}

\subsection{Boundary prediction errors}
\label{sec:boundary-errors}

{Without arity-indexed reductions, errors in the recovered child sequence are expressed through boundary-control decisions. The parser must determine where a reduction domain begins and whether its delimiter should be consumed or preserved. Table~\ref{tab:boundary-errors} distinguishes four errors: missing \(\textsc{D-Shift}\), where a required delimiter is not introduced; spurious \(\textsc{D-Shift}\), where an unnecessary delimiter is introduced; premature \(\textsc{Reduce}\), where a boundary is consumed before the highest constituent sharing it has been constructed; and spurious \(\textsc{D-Reduce}\), where a boundary is preserved after its final reduction domain has closed.}

\begin{table}
\centering
\small
\begin{tabular}{lcc}
\toprule
Error type & PTB & CTB \\
\midrule
Missing \(\textsc{D-Shift}\) & 369 & 102 \\
Spurious \(\textsc{D-Shift}\) & 293 & 77 \\
Premature \(\textsc{Reduce}\) & 455 & 142 \\
Spurious \(\textsc{D-Reduce}\) & 443 & 96 \\
\bottomrule
\end{tabular}
\caption{{Boundary-control errors in the delimiter-guided parser.}}
\label{tab:boundary-errors}
\end{table}

{Delimiter-persistence errors are more frequent than delimiter-insertion errors on both treebanks. The principal difficulty is therefore not only locating a left boundary but also determining how long it remains active across nested reductions. An incorrect recovered arity arises through these boundary decisions rather than through direct misclassification of an arity-indexed action.}

{Delimiter guidance thus relocates domain-size information from action labels to boundary control, reducing the action inventory at the cost of longer derivations and boundary-sensitive errors.}

\section{Conclusion}
\label{sec:conclusion}

Direct non-binary bottom-up composition does not require arity as an independent transition parameter. In every well-formed phrase-level reduction configuration, the active delimiter and label marker uniquely determine an ordered nonempty sequence of completed children. Reduction arity is therefore the cardinality of that sequence and follows from the configuration.

This recoverability yields a local correspondence with arity-specific reduction. Given a delimiter-bounded sequence \(s_1,\ldots,s_k\) and mother label \(X\), the delimiter-guided reduction and \(\textsc{Reduce-}X\#k\) construct the same constituent \(X(s_1,\ldots,s_k)\) after auxiliary control symbols are disregarded. The systems differ in where they encode the information identifying the reduction domain: the arity-specific system records its cardinality in the action label, whereas the delimiter-guided system records its boundary in the parser state and supplies the mother label separately.

The controlled PTB and CTB experiments support the empirical adequacy of this factorization. Delimiter guidance substantially reduces the action inventory, preserves the observed non-binary branching profile, and produces no systematic collapse at higher arities. Its operational cost is a longer derivation and boundary-sensitive errors involving delimiter insertion, preservation, and consumption. The essential object of bottom-up composition is thus the ordered reduction domain; once that domain is uniquely identified, arity is a derived property rather than a primitive decision.

The separation of mother-label prediction from reduction-domain identification suggests a unified delimiter-guided formulation of direct non-binary top-down, in-order, and bottom-up parsing, with traversal order determining when the mother label becomes available relative to its children; we leave the formal development of this generalization to future work.

\section*{Limitations}

The delimiter-guided factorization reduces the action inventory but lengthens derivations by separating mother-label prediction from constituent construction. It also replaces explicit arity errors with boundary-control errors, which may affect subsequent reductions that share the same delimiter.

The evaluation is limited to English PTB and Chinese CTB and assumes projective trees with contiguous constituent yields. Each system is evaluated with one fixed-seed run and greedy decoding under a shared Stanza-based architecture. The results therefore support a controlled comparison of transition representations but do not establish statistical superiority or generalization across languages and parser architectures.

\section*{Acknowledgments}
This work was supported by the Institute for Information \& Communications Technology Planning \& Evaluation (RS-2025-25461932, \textit{the Top-Tier AI Global HRD invitation} program, and RS-2022-II220078, \textit{Explainable Logical Reasoning for Medical Knowledge Generation}).

% % % Bibliography entries for the entire Anthology, followed by custom entries
% % %\bibliography{anthology,custom}
% % Custom bibliography entries only
% \bibliography{references}

\appendix

\section{Previous-work reimplementation}
\label{app:previous-work-reimplementation}

{We reimplement binary and arity-specific bottom-up baselines in the same
neural parsing framework as the delimiter-guided parser. All three systems share
the sentence encoder, parser-state representation, subtree-composition function,
action scorer, optimization procedure, decoder implementation, and evaluation
protocol. The two direct non-binary systems differ only in their transition
inventories, applicability conditions, and static oracles. The binary baseline
additionally replaces the original trees with binarized training targets.}

{A configuration consists of a stack \(\sigma\), the index \(i\) of the next
input token, and a completion flag \(f\). Stack items are shifted terminals or
completed subtrees. All nonfinal transitions require \(f=\mathit{false}\), and
\textsc{Finish} requires complete input consumption and a stack containing
exactly one completed root tree.}

\subsection{Binary bottom-up}

{The binary baseline is a bottom-up shift--reduce parser over deterministically
binarized trees \citep{sagae-lavie-2005-classifier}. \textsc{Shift} consumes the
next token, binary reductions combine the two topmost completed items, and unary
reduction projects a label over the topmost item. The left- and right-reduction
types record the headed binary structure prescribed by the binarization.}

\begin{figure*}
\centering
\small
\[
\begin{array}{ll}
\textsc{Shift} &
  \langle \sigma,\ i,\ \mathit{false} \rangle
  \Rightarrow
  \langle \sigma \mid w_i,\ i+1,\ \mathit{false} \rangle
\\[1ex]

\textsc{Reduce-Left-}X &
  \langle \sigma \mid C_1 \mid C_2,\ i,\ \mathit{false} \rangle
  \Rightarrow
  \langle \sigma \mid X(C_1,C_2),\ i,\ \mathit{false} \rangle
\\[1ex]

\textsc{Reduce-Right-}X &
  \langle \sigma \mid C_1 \mid C_2,\ i,\ \mathit{false} \rangle
  \Rightarrow
  \langle \sigma \mid X(C_1,C_2),\ i,\ \mathit{false} \rangle
\\[1ex]

\textsc{Reduce-Unary-}X &
  \langle \sigma \mid C,\ i,\ \mathit{false} \rangle
  \Rightarrow
  \langle \sigma \mid X(C),\ i,\ \mathit{false} \rangle
\\[1ex]

\textsc{Finish} &
  \langle T,\ n,\ \mathit{false} \rangle
  \Rightarrow
  \langle T,\ n,\ \mathit{true} \rangle .
\end{array}
\]
\caption{{Transitions of the binary bottom-up baseline; \(T\) is a completed
root tree.}}
\label{fig:binary-bottom-up-transitions}
\end{figure*}

{A constituent with more than two children is represented by a binary spine
containing intermediate nodes and is therefore constructed through several
reductions. Debinarization removes these auxiliary nodes and restores the
original non-binary constituent.}

\paragraph{Binarization for the binary baseline}

{The binary baseline is trained on deterministic headed binarizations of the
gold trees. Binarization replaces each non-binary node with an order-preserving
binary spine and labels introduced intermediate nodes with an
\(\texttt{@}\)-prefix. Punctuation remains in the ordered child sequence and is
treated as boundary structure rather than as a phrasal head. The transformation
is reversible: debinarization removes only the introduced intermediate nodes,
so evaluation is performed against the original treebank representation.}

\subsection{Arity-specific non-binary bottom-up}

{The second baseline reimplements the non-binary bottom-up system of
\citet{gonzalez-rodriguez-2019-faster}. It operates directly on the original
non-binary trees and uses an arity-specific reduction to combine the top \(k\)
completed stack items under a mother label \(X\).}

\begin{figure*}
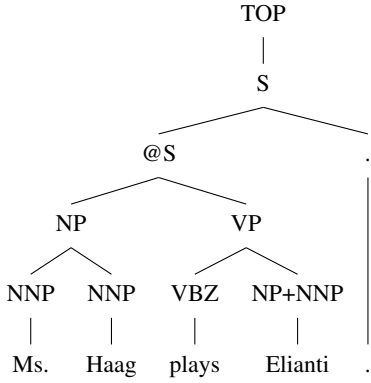

\centering
\small
\[
\begin{array}{ll}
\textsc{Shift} &
  \langle \sigma,\ i,\ \mathit{false} \rangle
  \Rightarrow
  \langle \sigma \mid w_i,\ i+1,\ \mathit{false} \rangle
\\[1ex]

\textsc{Reduce-}X\#k &
  \langle \sigma \mid C_1 \mid \cdots \mid C_k,\ i,\ \mathit{false} \rangle
  \Rightarrow
  \langle \sigma \mid X(C_1,\ldots,C_k),\ i,\ \mathit{false} \rangle
\\[1ex]

\textsc{Finish} &
  \langle T,\ n,\ \mathit{false} \rangle
  \Rightarrow
  \langle T,\ n,\ \mathit{true} \rangle .
\end{array}
\]
\caption{{Transitions of the arity-specific non-binary bottom-up baseline.}}
\label{fig:previous-nonbinary-bottom-up-transitions}
\end{figure*}

{The arity-specific and delimiter-guided systems construct the same original
non-binary target representation. Their difference is the location of
reduction-domain information: the arity-specific action supplies the number of
children directly, whereas the delimiter-guided system recovers the complete
child sequence from its configuration.}

\subsection{An example}
\label{subsec:baseline-example}

Figure~\ref{fig:oracle-example-binary-tree} shows the binarized representation used by the binary baseline, including the intermediate node \(\texttt{@S}\) and the collapsed unary-chain label \(\texttt{NP+NNP}\). Figure~\ref{fig:baseline-oracle-examples} gives the binary oracle sequence for this representation and the arity-specific oracle sequence for the original non-binary tree in Figure~\ref{fig:oracle-example-tree}.

\begin{figure}
\centering
\small
\begin{forest}
for tree={
  align=center,
  rounded corners,
  if n children=0{
    tier=word
  }{}
}
[TOP
  [S [@S
    [NP
      [NNP [Ms.]]
      [NNP [Haag]]
    ]
    [VP
      [VBZ [plays]]
      [NP+NNP [Elianti]]
    ] ]
    [. [.]]
  ]
]
\end{forest}
\caption{{Punctuation-aware binary representation of the example tree.}}
\label{fig:oracle-example-binary-tree}
\end{figure}

\begin{figure}
\centering
\footnotesize
\begin{tabular}{cll}
\toprule
& \multicolumn{1}{c}{\(\mathcal{B}\)}
& \multicolumn{1}{c}{\(\mathcal{A}\)} \\
\midrule
1  & \textsc{Shift}(\textit{Ms.})          & \textsc{Shift}(\textit{Ms.}) \\
2  & \textsc{Reduce-Unary}(\text{NNP})     & \textsc{Reduce-NNP}\#1 \\
3  & \textsc{Shift}(\textit{Haag})         & \textsc{Shift}(\textit{Haag}) \\
4  & \textsc{Reduce-Unary}(\text{NNP})     & \textsc{Reduce-NNP}\#1 \\
5  & \textsc{Reduce-Right}(\text{NP})      & \textsc{Reduce-NP}\#2 \\
6  & \textsc{Shift}(\textit{plays})        & \textsc{Shift}(\textit{plays}) \\
7  & \textsc{Reduce-Unary}(\text{VBZ})     & \textsc{Reduce-VBZ}\#1 \\
8  & \textsc{Shift}(\textit{Elianti})      & \textsc{Shift}(\textit{Elianti}) \\
9  & \textsc{Reduce-Unary}(\text{NP+NNP})  & \textsc{Reduce-NNP}\#1 \\
10 & \textsc{Reduce-Right}(\text{VP})      & \textsc{Reduce-NP}\#1 \\
11 & \textsc{Reduce-Right}(\text{@S})      & \textsc{Reduce-VP}\#2 \\
12 & \textsc{Shift}(\textit{.})            & \textsc{Shift}(\textit{.}) \\
13 & \textsc{Reduce-Unary}(\text{.})       & \textsc{Reduce-.}\#1 \\
14 & \textsc{Reduce-Right}(\text{S})       & \textsc{Reduce-S}\#3 \\
15 & \textsc{Reduce-Unary}(\text{TOP})     & \textsc{Reduce-TOP}\#1 \\
16 & \textsc{Finish}                       & \textsc{Finish} \\
\bottomrule
\end{tabular}
\caption{{Binary and arity-specific oracle sequences for the example tree.}}
\label{fig:baseline-oracle-examples}
\end{figure}

{The binary oracle constructs the ternary sentence constituent through the
intermediate \(\texttt{@S}\) node and two binary reductions. The arity-specific
oracle instead constructs the same constituent directly with a single ternary
reduction.}

\section{Experimental details}
\label{app:experimental-details}

All systems are implemented in the same Stanza constituency parsing framework \citep{bauer-manning-2025-high}.\footnote{\url{https://github.com/stanfordnlp/stanza/tree/main/stanza/models/constituency}} We retain the default neural configuration, except
that all models are trained for 100 epochs. The parser uses LSTM-based
transition and constituent stacks, a hidden size of 512, transition embeddings of
size 20, two LSTM layers, three output layers, ReLU nonlinearities, and max-based
constituent composition. Dropout settings are also fixed across systems, with
word dropout, prediction dropout, and LSTM input dropout set to 0.2.

Training uses the same optimization and batching settings for all systems. Unless
otherwise specified, we keep Stanza's default training configuration, including a
training batch size of 30, an evaluation batch size of 50, and the default
multistage optimization schedule. The train, development, and test splits are
identical across systems.

The systems differ only in transition inventory, static oracle, and the tree
transformation required by the transition system. The binary bottom-up baseline
is trained on binarized trees and evaluated after debinarization. The
arity-specific and delimiter-guided systems construct the original non-binary
trees directly. Sentence encoding, stack and buffer representations, subtree
composition, action scoring, optimization, decoding, and evaluation are held
fixed, so that the comparison isolates the effect of transition design.

\end{document}